\documentclass[sigconf]{acmart}
\renewcommand\footnotetextcopyrightpermission[1]{}

\usepackage{hyperref}

\usepackage{booktabs}
\usepackage[table,xcdraw]{xcolor}

\definecolor{strNeg}{RGB}{202,0,32}    
\definecolor{midNeg}{RGB}{244,165,130} 
\definecolor{neutral}{RGB}{247,247,247}
\definecolor{midPos}{RGB}{146,197,222}
\definecolor{strPos}{RGB}{5,113,176}   

\newcommand{\cNegS}[1]{\cellcolor{strNeg}\color{white}#1} 
\newcommand{\cNegM}[1]{\cellcolor{midNeg}#1}
\newcommand{\cNeu}[1]{\cellcolor{neutral}#1}
\newcommand{\cPosM}[1]{\cellcolor{midPos}#1}
\newcommand{\cPosS}[1]{\cellcolor{strPos}\color{white}#1} 

\AtBeginDocument{%
  }

\setcopyright{none}
\begin{document}
\fancyhead{}



\title{Beyond Symmetric Alignment: Spectral Diagnostics of Modality Imbalance in Vision-Language Models in the Medical Domain}

\author{Alessandro Gambetti}
\affiliation{%
  \institution{NOVA School of Science and Technology}
  \city{Lisbon}
  \country{Portugal}
}
\email{a.gambetti@campus.fct.unl.pt}

\author{Qiwei Han}
\affiliation{%
  \institution{Nova School of Business and Economics}
  \city{Lisbon}
  \country{Portugal}}
\email{qiwei.han@novasbe.pt}

\author{Cláudia Soares}
\affiliation{%
  \institution{NOVA School of Science and Technology}
  \city{Lisbon}
  \country{Portugal}
}
\email{claudia.soares@fct.unl.pt}

\author{Hong Shen}
\affiliation{%
 \institution{Carnegie Mellon University}
 \city{Pittsburgh}
 \state{Pennsylvania}
 \country{USA}}
\email{hongs@andrew.cmu.edu}

\renewcommand{\shortauthors}{Gambetti et al.}

\begin{abstract}
Vision-Language Models (VLMs) struggle when applied to medical image-text data, yet the tools available to diagnose this failure remain limited. 
Existing representation alignment metrics are symmetric, collapsing both modalities into a single score and hiding which modality drives cross-modal degradation. We introduce the Spectral Alignment Score (SAS), an asymmetric metric that projects both modalities onto the principal eigenbasis of an anchor modality and computes eigenvalue-weighted per-eigenmode correlations, resulting in directional scores whose difference quantifies modality information imbalance. 
We embed SAS within a benchmarking framework evaluating 15 VLMs across natural and medical image-text datasets alongside 6 alignment metrics and bidirectional retrieval. 
Our experiments show that medical images retain richer structural information than their paired clinical reports, a directional asymmetry invisible to all competing metrics, and that SAS achieves the strongest zero-label correlation with retrieval performance in the medical domain, positioning it as a practical diagnostic tool for clinical deployment. 
Code is available at this {\color{blue} https://github.com/iamalegambetti/medical-vlms-assessment}.
\end{abstract}




\maketitle

\section{Introduction}

AI has been increasingly integrated into clinical workflows to support decision systems, including diagnosis, treatment planning, and patient outcome assessment, with evidence suggesting that well-designed systems can augment clinician performance in high-stakes scenarios~\cite{sendak2020human,longoni2019resistance,zhang2023ethics,ghassemi2021false}. 
%
%
Currently, Vision-Language Models (VLMs) have been adopted in medical applications that involve image-text data to pair radiological images to relevant reports without requiring manually annotated pairs at inference time, a meaningful advantage in clinical settings where expert annotation is scarce and expensive~\cite{hartsock2024vision}. 
The dominant training paradigm for VLMs combines dual-encoder architectures with contrastive objectives, optimizing for a shared representational space in which semantically aligned image-text pairs are mapped together while negative pairs are pushed apart~\cite{oord2018representation,wang2020understanding,radford2021learning}. Under this formulation, visual and textual encoders are jointly trained to maximize mutual information across modalities, yielding transferable representations that generalize across distribution shifts without task-specific supervision. 
Models such as CLIP~\cite{radford2021learning}, MetaCLIP~\cite{xu2024demystifying}, SigLIP~\cite{zhai2023sigmoid}, and SigLIP 2~\cite{tschannen2025siglip2} have demonstrated strong zero-shot transfer across a broad range of visual recognition benchmarks and are extensively deployed as feature extractors within larger multimodal systems~\cite{liu2023visual,alayrac2022flamingo}.
However, despite this versatility, a critical gap emerges when general-purpose VLMs are deployed on medical image-text data. These models are generally pre-trained on web-crawled natural image-caption pairs, whose characteristics differ substantially from radiology data, where images are highly information-dense, and reports are often concise, domain-specific, and structurally different from web text.
This distributional mismatch induces an alignment failure,
which is well-documented in natural settings but poorly characterized in the medical domain. 
The tools available for measuring this misalignment are many, but often sub-optimal for the task. Existing representation alignment metrics, including CKA~\cite{kornblith2019similarity}, SVCCA [17], CORAL~\cite{sun2016deep}, MMD~\cite{gretton2012kernel}, and RMG~\cite{schrodi2025TwoEO}, are all symmetric: they reduce cross-modal alignment to a single bidirectional score, and are therefore blind to which modality drives cross-modal degradation. In medical imaging, where visual structure is far richer than what paired clinical reports often express, this directional asymmetry is precisely what matters and what existing metrics cannot measure.

In this paper, we introduce the \textbf{Spectral Alignment Score (SAS)}, a family of metrics that address this gap directly. SAS scores can be used both symmetrically and asymmetrically, projecting both modalities onto the principal eigenbasis of an anchor modality and computing eigenvalue-weighted per-eigenmode correlations over the top spectral directions, yielding two directional scores \textit{i.e.}, $S_{X \to Y}$ and $S_{Y \to X}$, whose difference $\Delta_{SAS}$ quantifies modality information imbalance. 
To this end, we employ SAS alongside existing representation alignment metrics within a benchmarking framework that evaluates 15 VLMs, 13 general-purpose and 2 medical, across natural (MS-COCO 2014, Flickr30k) and medical (ROCO, MIMIC-CXR) datasets. 
Concretely, we conduct four experiments: 
(1) we measure domain shift in cross-modal alignment across datasets, showing that medical datasets are associated with less meaningful representations; 
(2) we characterize the directional modality imbalance via $\Delta_{SAS}$, in which the dominant image eigenspace is significantly more recoverable from text than the reverse; 
(3) we sweep the eigenspectrum depth parameter $q$ to assess its sensitivity to the representations rank, providing evidence that most of the information is concentrated in the very first dimensions; 
and finally (4) we evaluate how well alignment metrics predict downstream retrieval performance, showing that SAS is a robust retrieval predictor.
%
%
%
Our contributions can be summarized as follows.
\begin{enumerate}
    \item We introduce SAS,
    a metric that decomposes cross-modal alignment into two
    directed eigenspace-recovery scores, $S_{X \to Y}$ and $S_{Y \to X}$,
    and their difference $\Delta_{\mathrm{SAS}}$. SAS is the first alignment
    metric for dual-encoder VLMs that can be spectrally deployed both asymmetrically and symmetrically.

    \item We conduct a systematic benchmark evaluating 15~VLMs across four
    image-text datasets~(MS-COCO~2014, Flickr30k, ROCO, MIMIC-CXR) using
    six alignment metrics alongside SAS and bidirectional retrieval, providing
    a unified diagnostic comparison across natural and medical domains.

    \item We identify a consistent directional asymmetry in the medical
    domain: across all 13 general-purpose VLMs and both medical datasets,
    the dominant image eigenspace is significantly more recoverable from text
    embeddings than the reverse~($\Delta_{\mathrm{SAS}} = +0.049$,
    $p < 0.001$), while not being statistically different from zero
    on natural data. This asymmetry is invisible to all symmetric metrics.

    \item We provide evidence that cosine similarity and SAS are
    complementary diagnostic metrics, with the former measuring whether paired
    and unpaired items are separable in cosine space, and the latter identifying which modality drives failure.
    Together they form a label-free pre-deployment audit that requires no
    annotated pairs.
\end{enumerate}
In summary, SAS adds a missing diagnostic dimension to multimodal evaluation, which is not meant to replace but to be used in synergy with current evaluation metrics.

\section{Representation Alignment Metrics}
To evaluate the alignment between learned representations, we first define the mathematical notation and then review the metrics considered in this study. 
Let $X \in \mathbb{R}^{n \times d_1}$ and $Y \in \mathbb{R}^{n \times d_2}$ denote representation matrices from two neural network layers or modalities, in our case images and text, respectively, where $n$ represents the number of data points (samples), and $d_1, d_2$ denote the respective feature dimensions. Let $H = I_n - \frac{1}{n}\mathbf{1}\mathbf{1}^T$ be the centering matrix, ensuring that the column-centered representations are given by $\tilde{X} = HX$ and $\tilde{Y} = HY$. 

\subsection{Singular Vector Canonical Correlation Analysis (SVCCA)~\cite{raghu2017svcca}}
SVCCA combines Singular Value Decomposition (SVD) and Canonical Correlation Analysis (CCA) to handle over-parameterized layers efficiently \cite{raghu2017svcca}. Given $\tilde{X}$ and $\tilde{Y}$, SVD is first applied to filter out noise and retain the top directions accounting for $99\%$ of the variance, yielding low-rank representations $\hat{X} \in \mathbb{R}^{n \times k_1}$ and $\hat{Y} \in \mathbb{R}^{n \times k_2}$. Standard CCA is then performed on $\hat{X}$ and $\hat{Y}$ by finding projection vectors $w_X^{(i)}$ and $w_Y^{(i)}$ that maximize the CC $\rho_i$:
\begin{equation}
    \rho_i = \max_{w_X^{(i)}, w_Y^{(i)}} \frac{(w_X^{(i)})^T \hat{X}^T \hat{Y} w_Y^{(i)}}{\sqrt{(w_X^{(i)})^T \hat{X}^T \hat{X} w_X^{(i)} \cdot (w_Y^{(i)})^T \hat{Y}^T \hat{Y} w_Y^{(i)}}}
\end{equation}
subject to orthogonality with prior directions. The final SVCCA similarity is the mean of the CC: $\bar{\rho} = \frac{1}{\min(k_1, k_2)} \sum_{i=1}^{\min(k_1, k_2)} \rho_i$.

\subsection{Centered Kernel Alignment (CKA)~\cite{kornblith2019similarity}}
While CCA-based methods are invariant to any invertible linear transformation, they can fail to identify correspondences when $d_1, d_2 > n$. CKA relaxes this by offering invariance only to orthogonal transformations and isotropic scaling. Let $K = \kappa(X, X)$ and $L = \ell(Y, Y)$ be the $n \times n$ similarity matrices computed via semidefinite kernels $\kappa$ and $\ell$. Using the Hilbert-Schmidt Independence Criterion (HSIC), CKA is defined as:
\begin{equation}
    \text{CKA}(K, L) = \frac{\text{HSIC}(K, L)}{\sqrt{\text{HSIC}(K, K) \text{HSIC}(L, L)}}
\end{equation}
where $\text{HSIC}(K, L) = \frac{1}{(n-1)^2} \text{Tr}(HKHL)$. In the linear case where $K = XX^T$ and $L = YY^T$, this simplifies using the Frobenius norm $\|\cdot\|_F$:
\begin{equation}
    \text{Linear CKA}(X, Y) = \frac{\|\tilde{Y}^T \tilde{X}\|_F^2}{\|\tilde{X}^T \tilde{X}\|_F \|\tilde{Y}^T \tilde{Y}\|_F}
\end{equation}

\subsection{Correlation Alignment (CORAL)~\cite{sun2016deep}}
CORAL addresses unsupervised domain adaptation by directly minimizing the discrepancy between the second-order statistics of a source domain $X$ and a target domain $Y$ \cite{sun2016return}. The feature covariance matrices are defined as $\Sigma_X = \frac{1}{n-1} X^T H X$ and $\Sigma_Y = \frac{1}{n-1} Y^T H Y$. CORAL acts as a misalignment metric function penalizing the distance between these covariances:
\begin{equation}
    \text{CORAL}= \frac{1}{4 d^2} \|\Sigma_X - \Sigma_Y\|_F^2
\end{equation}
where $d = d_1 = d_2$. 

\subsection{Maximum Mean Discrepancy (MMD)~\cite{gretton2012kernel}}
MMD is a non-parametric kernel-based metric used to determine whether two samples are drawn from different distributions \cite{gretton2012kernel}. It measures the distance between the embeddings of probability distributions $p$ and $q$ within a Reproducing Kernel Hilbert Space (RKHS) $\mathcal{H}$. Given a kernel function $k(\cdot,\cdot)$ associated with feature map $\phi$, the squared MMD between sample batches $X \sim p$ and $Y \sim q$ is given by:
\begin{equation}
    \text{MMD}^2(X, Y) = \frac{1}{n^2} \sum_{i,j=1}^n k(x_i, x_j) - \frac{2}{n^2} \sum_{i,j=1}^n k(x_i, y_j) + \frac{1}{n^2} \sum_{i,j=1}^n k(y_i, y_j)
\end{equation}
An MMD score of $0$ indicates perfect alignment between the two representation distributions.

\subsection{Relative Modality Gap (RMG)~\cite{schrodi2025TwoEO}}
In contrastive vision-language models (\textit{e.g.}, CLIP), a systematic geometric shift separates image features $X$ and text features $Y$ in the shared embedding space. To measure this misalignment while accounting for cross-modal structural density, the Relative Modality Gap (RMG) normalizes the absolute modality gap vector by the internal variance of the representations:
\begin{equation}
    \text{RMG}(X, Y) = \frac{\| \mu_X - \mu_Y \|_2}{\frac{1}{2} ( \sigma_X + \sigma_Y )}
\end{equation}
where $\mu_X = \frac{1}{n}\sum_{i=1}^n x_i$ and $\mu_Y = \frac{1}{n}\sum_{i=1}^n y_i$ are the centroids of each modality, and $\sigma_X = \sqrt{\frac{1}{n}\sum_{i=1}^n \|x_i - \mu_X\|_2^2}$ (and analogously $\sigma_Y$ for $Y$) represents the average standard deviation within each respective modality space. Higher values indicate higher modality gap.

\section{Spectral Alignment Score (SAS)}
We introduce Spectral Alignment Score (SAS) to measure both alignment and directional information imbalance.
Assuming equal feature dimensions ($d_1 = d_2 = d$), we first compute the empirical covariance of the centered features $\tilde{X}$, denoted as $\Sigma_X = \frac{1}{n} \tilde{X}^T \tilde{X}$, and its eigendecomposition $\Sigma_X = V \Lambda V^T$, where $\Lambda = \mathrm{diag}(\lambda_1, \dots, \lambda_d)$ with $\lambda_1 \ge \cdots \ge \lambda_d \ge 0$. 

We project both modalities onto the principal directions of $X$, yielding $Z_X = \tilde{X}V$ and $Z_Y = \tilde{Y}V$. The cross-covariance in this eigenbasis is $A = \frac{1}{n} Z_X^T Z_Y$, where the diagonal element $a_k = A_{kk}$ quantifies the alignment along the $k$-th eigenmode. Letting $\sigma_{Y,k}^2$ be the variance of the $k$-th column of $Z_Y$ and $\varepsilon$ a small stabilizing constant, the normalized correlation coefficient is given by:
\begin{equation} \label{eq:sas_corrcoef}
    \rho_k = \frac{a_k}{\sqrt{\lambda_k \, \sigma_{Y,k}^2 + \varepsilon}}
\end{equation}
To filter out noise and focus on the most informative directions, we define an active index set $\mathcal{I}_q = \{k : \lambda_k \ge \phi\}$, where $\phi$ is the $(1-q)$-quantile of the eigenvalues for $q \in (0,1]$. The asymmetric directional alignment score from $X$ to $Y$ is then computed as the eigenvalue-weighted sum of these correlations:
\begin{equation}
    S_{X \to Y} = \frac{\sum_{k \in \mathcal{I}_q} \lambda_k |\rho_k|}{\sum_{k \in \mathcal{I}_q} \lambda_k}
\end{equation}
%
%
SAS is a \emph{global} metric. 
Asymmetrically, $S_{X \to Y}$ measures, along $X$'s dominant principal directions (weighted by their explanatory variance $\lambda_k$), how strongly $Y$ covaries with $X$. Equivalently, $S_{Y \to X}$ is high when $X$'s dominant eigenspace is \emph{recoverable from} $Y$'s structure along the same directions. 
Symmetrically, SAS reduces as a weighted average: $\mathrm{SAS}(X, Y) = \frac{1}{2}(S_{X \to Y} + S_{Y \to X})$.
We briefly describe its properties:
\begin{itemize}
\item \emph{(Directionality).} $S_{X \to Y}$ and $S_{Y \to X}$ are not equal in general. $\Delta_{SAS} = S_{X \to Y} - S_{Y \to X} > 0$ indicates that $X$'s dominant eigenspace is more strongly recovered from $Y$ than $Y$'s dominant eigenspace is from $X$.
\item \emph{(Symmetric limit)}. If $X$ and $Y$ share identical covariance structure up to an orthogonal transformation and have comparable per-mode correlations, then $S_{X \to Y} \approx S_{Y \to X}$ and $\Delta_{\mathrm{SAS}} \approx 0$.
\item \emph{(Noise-tail dilution)}. Increasing $q$ admits lower-variance eigenmodes into $\mathcal{I}_q$. If these tail modes are weakly cross-aligned, SAS decreases or stabilises.
\item \emph{(Orthogonal invariance)}. SAS is invariant to orthogonal transformations applied within either modality.
\end{itemize}
SAS therefore exposes structural asymmetry between modalities that symmetric metrics (CKA, SVCCA, CORAL, MMD, RMG) average away.

\subsection{Summary of Metric Properties}

\begin{table*}[t]
\caption{Comparison of representation alignment metrics.
\textbf{Invariance definitions}: \textit{Affine} implies robustness to invertible
linear transformations and translation; \textit{Orthogonal} denotes robustness to
rotation and reflection; \textit{Translation} refers to robustness under constant
additive shifts.
\textbf{Scope}: \textit{Global} requires the full population matrix;
\textit{Local$^\dagger$} admits a per-instance score given a fixed population prior.
\textbf{Ordinal}: whether the score supports rank ordering with meaningful magnitude
within a fixed dataset and dimensionality.}
\label{tab:metric_comparison}
\centering
\small
\begin{tabular}{@{}llccp{10cm}@{}}
\toprule
\textbf{Metric} & \textbf{Invariance} & \textbf{Asymmetric} & \textbf{Ordinal} & \textbf{Scope / Focus} \\ \midrule
\textbf{SVCCA} \cite{raghu2017svcca}
  & Affine & No & Yes ($\uparrow$, $[0,1]$) & Global. Measures the mean canonical correlation between the top SVD subspaces of two representations, retaining directions that explain 99\% of variance. \\[4pt]
\textbf{CKA} \cite{kornblith2019similarity}
  & Orthogonal & No & Yes ($\uparrow$, $[0,1]$) & Global. Quantifies the normalised HSIC between pairwise similarity structures, comparing the relational geometry of two representation sets rather than their individual directions. \\[4pt]
\textbf{CORAL} \cite{sun2016return}
  & None & No & Yes ($\downarrow$, $[0,\infty)$) & Global. Measures the squared Frobenius distance between the full second-order covariance matrices of two distributions, penalising any mismatch in spread and correlation structure. \\[4pt]
\textbf{MMD} \cite{gretton2012kernel}
  & None & No & Yes ($\downarrow$, $[0,\infty)$) & Global. Estimates the squared distance between kernel mean embeddings of two distributions in an RKHS, functioning as a non-parametric two-sample test statistic. \\[4pt]
\textbf{RMG} \cite{schrodi2025TwoEO}
  & Translation & No & Yes ($\downarrow$, $[0,1]$) & Global$^\ddagger$. Normalises the mean paired cosine distance between modalities by the average intra-modal spread, isolating the geometric separation of the two representation clouds. \\ \midrule
\textbf{SAS} (Ours)
  & Orthogonal & Yes & Yes ($\uparrow$, $[0,1]$) & Global$^\dagger$. Projects both modalities onto the anchor eigenbasis and computes eigenvalue-weighted per-eigenmode correlations, exposing which modality retains more structural information in the dominant spectral directions. Can be decomposed into its symmetric components: $S_{X\to Y}$ and $S_{Y \to X}$. \\ \bottomrule
\end{tabular}
\begin{flushleft}
\footnotesize
$^\dagger$SAS admits per-instance scoring by inheriting the population eigenbasis,
eigenvalues, and per-eigenmode $Y$-variance as a fixed prior; only the rank-1
cross-covariance diagonal is instance-specific.\\
$^\ddagger$RMG's numerator averages per-pair cosine distances, making individual pair
contributions well-defined, but the denominator (intra-modal mean distance) remains
a global statistic.
\end{flushleft}
\end{table*}

As summarized in Table~\ref{tab:metric_comparison}, existing representation alignment metrics predominantly offer symmetric, bidirectional assessments. Classical methods like SVCCA and CKA provide rigorous invariance profiles (such as affine or orthogonal invariance) but collapse the structural characteristics of both spaces into a single omnidirectional score. While they are highly effective for determining whether two architectures have converged to identical global representations, they fail to diagnose scenarios where one modality is less informative than the other.

SAS fundamentally diverges from this paradigm by preserving directionality through its asymmetric formulations ($S_{X \to Y}$ and $S_{Y \to X}$). By projecting features into the specific eigenbasis of an anchor representation ($X$) and weighting the localized correlations ($\rho_k$) by the explicit explanatory variance ($\lambda_k$), SAS naturally exposes the \textit{information imbalance} between distinct representation modalities. Unlike CORAL or linear CKA, which aggregate all second-order statistics indiscriminately, SAS incorporates a robust noise-filtering mechanism via the quantile index set $\mathcal{I}_q$. This makes it uniquely suitable for analyzing multimodal spaces asymmetrically.

\section{Experiments}

\begin{figure*}[t]
    \centering
    \includegraphics[width=0.85\linewidth]{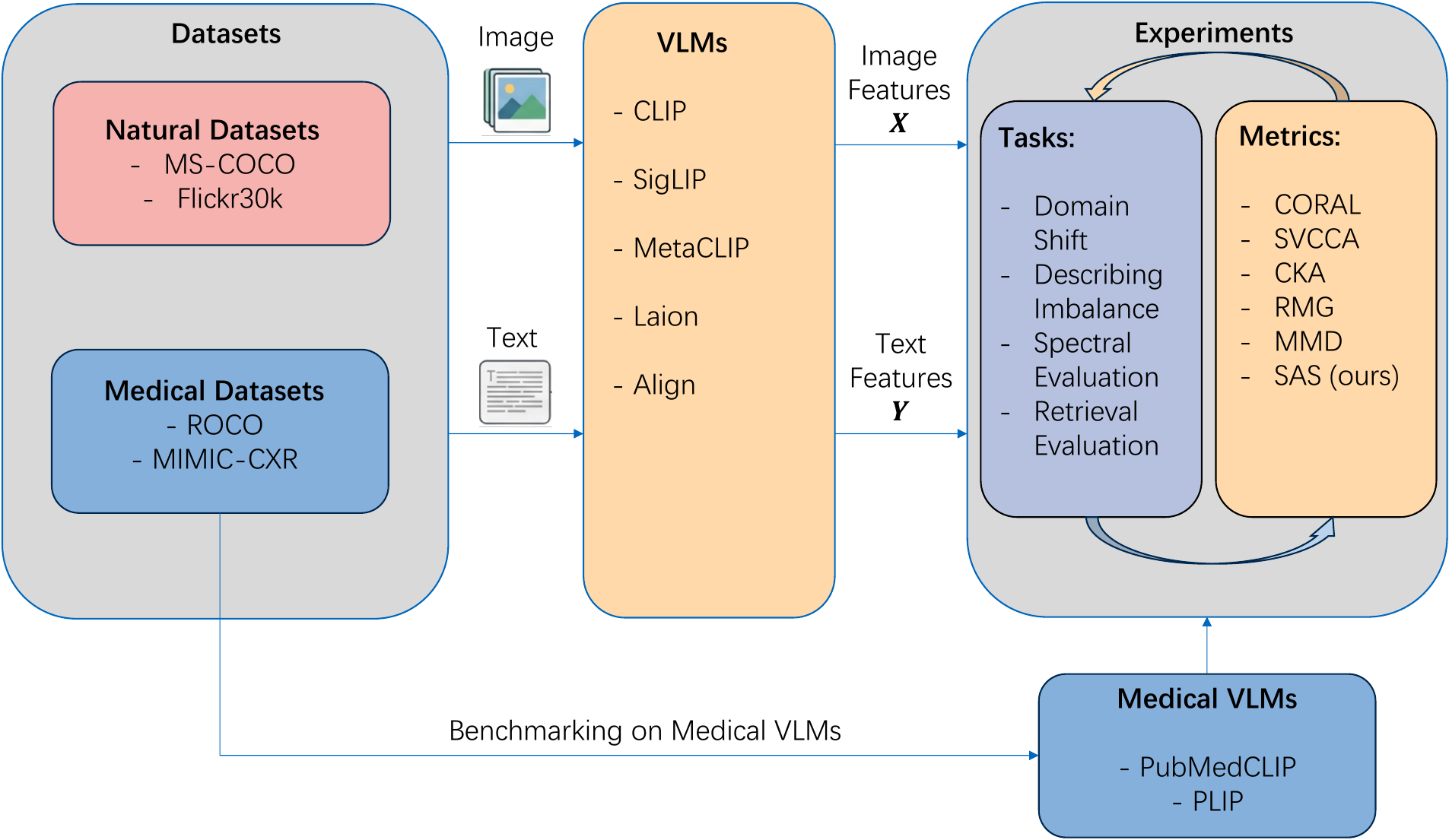}
    \caption{Summary of our proposed framework, which uses natural datasets (MS-COCO 2014, Flickr30k) and medical datasets (ROCO, MIMIC-CXR) to extract image and text features. It tests general VLMs (CLIP, SigLIP, MetaCLIP) and medical VLMs (PubMedCLIP, PLIP) across several tasks, including retrieval, domain shift imbalance, and spectral evaluations. This evaluation uses several representation learning metrics, including our proposed SAS.}\label{fig:methodology}
\end{figure*}

Figure~\ref{fig:methodology} illustrates our methodological approach. 
We briefly review the datasets and VLMs used. Then, for each experiment, we present the research design and discuss the related results.

\subsection{Datasets}
\textit{Medical}-domain datasets are ROCO~\cite{pelka2018roco} and MIMIC-CXR~\cite{johnson2019mimic}.
Both datasets include paired image-captions from Electronic Health Records (EHR) of several medical imaging modalities, including Computer Tomography, Ultrasound, X-Ray, Fluoroscopy, Positron Emission Tomography, Mammography, Magnetic Resonance Imaging, and Angiography. 
We benchmark our approach with standard image-text datasets such as MS-COCO 2014 and Flickr30k, which we call \textit{Natural} datasets~\cite{karpathy2015deep}.

\subsection{Pre-trained VLMs Zoo}

We adopt 13 open-source dual-encoder VLMs spanning from $\sim$150M to over 1B parameters. We include contrastive-learning models, \textit{e.g.}, CLIP~\cite{radford2021learning}, Align~\cite{jia2021scaling},MetaCLIP~\cite{xu2024demystifying}, MetaCLIP2~\cite{chuang2025meta}, and LAION-CLIP~\cite{schuhmann2022laionb} (built on OpenCLIP~\cite{cherti2023openclip}), as well as sigmoid-loss variants SigLIP~\cite{zhai2023sigmoid} and SigLIP2~\cite{tschannen2025siglip2}. For each model, we consider a base and a large version. We summarize models used in Table~\ref{tab:vlm_list}.
We also benchmark with PubMedCLIP~\cite{eslami2021does} and PLIP~\cite{huang2023visual}, two VLMs specifically designed for medical AI.

\begin{table}[t]
\centering
\caption{Summary of VLMs used in our study, including HuggingFace identifier and number of parameters.} \label{tab:vlms_summary}
\label{tab:vlm_list}
\small
\begin{tabular}{lc}
\toprule
\textbf{VLM (HF Identifier)} & \textbf{\#Params} \\
\midrule
\texttt{openai/clip-vit-base-patch32}~\cite{radford2021learning}        & $\sim$151M \\
\texttt{openai/clip-vit-large-patch14}~\cite{radford2021learning}       & $\sim$428M \\
\texttt{google/siglip-base-patch16-224}~\cite{zhai2023sigmoid}          & $\sim$200M \\
\texttt{google/siglip-so400m-patch14-384}~\cite{zhai2023sigmoid}        & $\sim$400M \\
\texttt{google/siglip2-base-patch16-224}~\cite{tschannen2025siglip2}    & $\sim$200M \\
\texttt{google/siglip2-so400m-patch14-384}~\cite{tschannen2025siglip2}  & $\sim$400M \\
\texttt{kakaobrain/align-base}~\cite{jia2021scaling}                    & $\sim$300M \\
\texttt{facebook/metaclip-b32-400m}~\cite{xu2024demystifying}           & $\sim$150M \\
\texttt{facebook/metaclip-h14-fullcc2.5b}~\cite{chuang2025meta}         & $\sim$1B+  \\
\texttt{facebook/metaclip-2-mt5-worldwide-b32}~\cite{chuang2025meta}    & $\sim$300M \\
\texttt{facebook/metaclip-2-worldwide-huge-378}~\cite{xu2024demystifying}     & $\sim$1B+  \\
\texttt{laion/CLIP-ViT-B-32-laion2B-s34B-b79K}~\cite{schuhmann2022laionb} & $\sim$150M \\
\texttt{laion/CLIP-ViT-L-14-laion2B-s32B-b82K}~\cite{schuhmann2022laionb} & $\sim$400M \\
\bottomrule
\end{tabular}
\end{table}

\subsection{Experimental Setup}
We further define the following metrics: 
$\Delta_{cos} = \frac{1}{n} \sum_{i=1}^{n} \cos(x_i, y_i) - \frac{1}{n(n-1)} \sum_{i \neq j} \cos(x_i, y_j)$
as the difference between the matched image-text pairs and the unmatched pairs (off-diagonal), with 0 meaning that the model cannot separate paired from unpaired items; and (2) $\Delta_{SAS} = S_{X \to Y} - S_{Y \to X}$ capturing the imbalance between the two modalities representation, positive when images ($X$) structure is better captured by text ($Y$), and vice versa. Naturally, a value of 0 implies perfect information sharing between modalities. 

To test for statistical significance, we employ the Mann-Whitney U test (MWU), both two-sided and one-sided, one-sample t-tests, and two-sided Spearman rank correlation tests. 
Statistical significance is indicated as *$ p < .05$, **$p < .01$, ***$p < .001$.


\subsection{Exp. 1 — Alignment Degrades Under Medical Domain Shift}

\subsubsection{Design}

We assess whether the metrics consistently detect alignment degradation when transitioning from natural to medical image-text datasets. The hypothesis is that, across all VLMs, all metrics 
have lower cross-modal alignment on ROCO and MIMIC-CXR. 
We compute the complete metric suite 
across all 52 general-purpose \texttt{(model $\times$ dataset)} pairs, totaling 26 natural and 26 medical observations per metric. 
Domain-level differences are assessed via two-sided MWU tests. 

\subsubsection{Results}

Table~\ref{tab:exp1_domain_summary} reports the mean and standard deviation of each metric across natural and medical domains for general-purpose models, together with MWU $p$-values.

\begin{table}[t]
\centering
\caption{Summary of domain-level alignment metrics aggregated per \texttt{(model x dataset)} pairs. $\uparrow$ denotes higher-is-better; $\downarrow$ lower-is-better.
}
\label{tab:exp1_domain_summary}
\begin{tabular}{lcrrrl}
\toprule
\textbf{Metric} & \textbf{Dir.} & \textbf{Natural} ($\mu \pm \sigma$) & \textbf{Medical} ($\mu \pm \sigma$) & \textbf{MWU} \\
\midrule
CKA                    & $\uparrow$ & $0.4590 \pm 0.0566$ & $0.2026 \pm 0.1737$ & $618$*** \\
CORAL                  & $\downarrow$ & $0.0776 \pm 0.0094$ & $0.0572 \pm 0.0169$ & $570$*** \\
RMG                    & $\downarrow$ & $0.7336 \pm 0.0332$ & $0.7213 \pm 0.0295$ & $412$ \\
$\Delta_{\cos}$        & $\uparrow$ & $0.2023 \pm 0.0323$ & $0.0386 \pm 0.0400$ & $676$*** \\
$S_{X{\to}Y}$         & $\uparrow$ & $0.5981 \pm 0.0302$ & $0.2800 \pm 0.2204$ & $662$*** \\
$S_{Y{\to}X}$         & $\uparrow$ & $0.6059 \pm 0.0372$ & $0.2314 \pm 0.1880$ & $676$*** \\
SAS     & $\uparrow$ & $0.6020 \pm 0.0293$ & $0.2557 \pm 0.2040$ & $676$*** \\
$\Delta_{\text{SAS}}$  & $0$      & $-0.0079 \pm 0.0338$ & $0.0486 \pm 0.0375$ & $98$*** \\
MMD                    & $\downarrow$ & $0.7478 \pm 0.1963$ & $1.0757 \pm 0.2388$ & $104$*** \\
SVCCA                  & $\uparrow$ & $0.4516 \pm 0.0487$ & $0.2871 \pm 0.1812$ & $495$** \\
\bottomrule
\end{tabular}
\end{table}

Nine of ten metrics detect the domain shift at $p < 0.05$; eight of these reach $p < 0.001$, with SVCCA being significant at $p = 0.01$. RMG is the sole exception ($p = 0.179$), consistent with its sensitivity to first-order modality gap rather than distributional structure~\cite{liang2022mind,schrodi2025TwoEO}.


The asymmetric structure of $\Delta_{\text{SAS}}$ shows a finding invisible to all other metrics. In the natural domain, $\Delta_{\text{SAS}} = -0.0079$ indicates approximate bidirectional balance, in which image and text eigenbases are symmetrically recoverable. However, in the medical domain, $\Delta_{\text{SAS}}$ rises to $+0.049$, implying that text representations better recover image structure than vice versa. This directional imbalance is invisible to symmetric metrics such as CKA or SVCCA.


\subsection{Exp. 2 — Medical Data Reveals a Consistent Image-over-Text Directional Asymmetry}

\subsubsection{Design}

We statistically test the directional asymmetry $\Delta_{\text{SAS}}$ in natural datasets and in medical datasets. The hypothesis is that natural datasets, constructed from balanced web-crawled corpora, present symmetric alignment in the dominant eigenmodes. In contrast, radiology data should show a positive imbalance because the visual structure of medical images may be more information-rich than their paired reports.
Directional SAS ($S_{X\to Y}$, $S_{Y\to X}$) is computed at $q = 0.1$ for all the 52 \texttt{(model $\times$ dataset)} pairs. Statistical significance is assessed via one-sample $t$-tests against $\mu = 0$ per domain.


\subsubsection{Results}

Table~\ref{tab:exp2_domain_imbalance} summarises $\Delta_{\text{SAS}}$ at the domain level for general-purpose models, and Table~\ref{tab:exp2_per_dataset_imbalance} breaks results down by individual dataset.

\begin{table}[t]
\centering
\caption{Domain-level $\Delta_{\text{SAS}}$ aggregated per \texttt{(model x dataset)} pairs. 
}
\label{tab:exp2_domain_imbalance}
\begin{tabular}{lrrl}
\toprule
\textbf{Domain} & $\mathbf{\Delta_{SAS}}$ ($\mu \pm \sigma$) & $\mathbf{N}$ & \textbf{$t$-test} ($\mu{=}0$) \\
\midrule
Natural & $-0.0079 \pm 0.0338$ & 26 & $-1.162$ \\
Medical & $+0.0486 \pm 0.0375$ & 26 & $ 6.486$*** \\
\bottomrule
\end{tabular}
\end{table}

\begin{table}[t]
\centering
\caption{$\Delta_{\text{SAS}}$  aggregated per \texttt{model} for every dataset. ``All > 0?'' indicates whether every model yields $\Delta_{\text{SAS}} > 0$.}
\label{tab:exp2_per_dataset_imbalance}
\begin{tabular}{llrr}
\toprule
\textbf{Dataset} & \textbf{Domain} & $\mathbf{\Delta_{sas}}$ ($\mu \pm \sigma$) & \textbf{All > 0?} \\
\midrule
MS-COCO 2014 & Natural & $-0.0058 \pm 0.0353$ & No \\
Flickr30k     & Natural & $-0.0099 \pm 0.0321$ & No \\
ROCO          & Medical & $+0.0819 \pm 0.0222$ & Yes \\
MIMIC-CXR     & Medical & $+0.0153 \pm 0.0099$ & Yes \\
\bottomrule
\end{tabular}
\end{table}

The results confirm the directional hypothesis. On natural data, $\Delta_{\text{SAS}}$ is not significantly different from zero ($-1.162$), indicating that image and text eigenbases are symmetrically recoverable from each other.
On the contrary, on medical data, $\Delta_{\text{SAS}}$ is significantly positive ($6.486$***).
At the per-dataset level, ROCO yields $\Delta_{\text{SAS}} > 0$ for every VLM, while MIMIC-CXR shows a smaller absolute imbalance, but with a positive sign.

The asymmetry is directionally consistent with the hypothesis: the image modality retains more structural information in the principal eigenmodes than the text side can recover, a pattern that symmetric metrics such as CKA or SVCCA cannot distinguish from the symmetric case of jointly degraded alignment.
This demonstrates that $\Delta_{\text{SAS}}$ can measure modality information imbalance.



\subsection{Exp. 3 — Top Eigenmodes Drive the Natural-Medical Gap}

\subsubsection{Design}

We investigate how the SAS sensitivity to domain shift varies as a function of the eigenspectrum depth parameter $q$. Specifically, we ask whether the natural-medical alignment gap is maximised at small $q$, consistent with the hypothesis that medical cross-modal alignment is concentrated in fewer top eigenmodes. We sweep $q \in \{0.1, 0.3, 0.5, 0.7, 1.0\}$ and compute $S_{X{\to}Y}$, $S_{Y{\to}X}$ and $\Delta_{\text{SAS}}$ for all the 52 \texttt{(model $\times$ dataset)} pairs, aggregating results by domain at each $q$ value. 

\subsubsection{Results}

\begin{figure*}[t]
    \centering
    \includegraphics[width=1\linewidth]{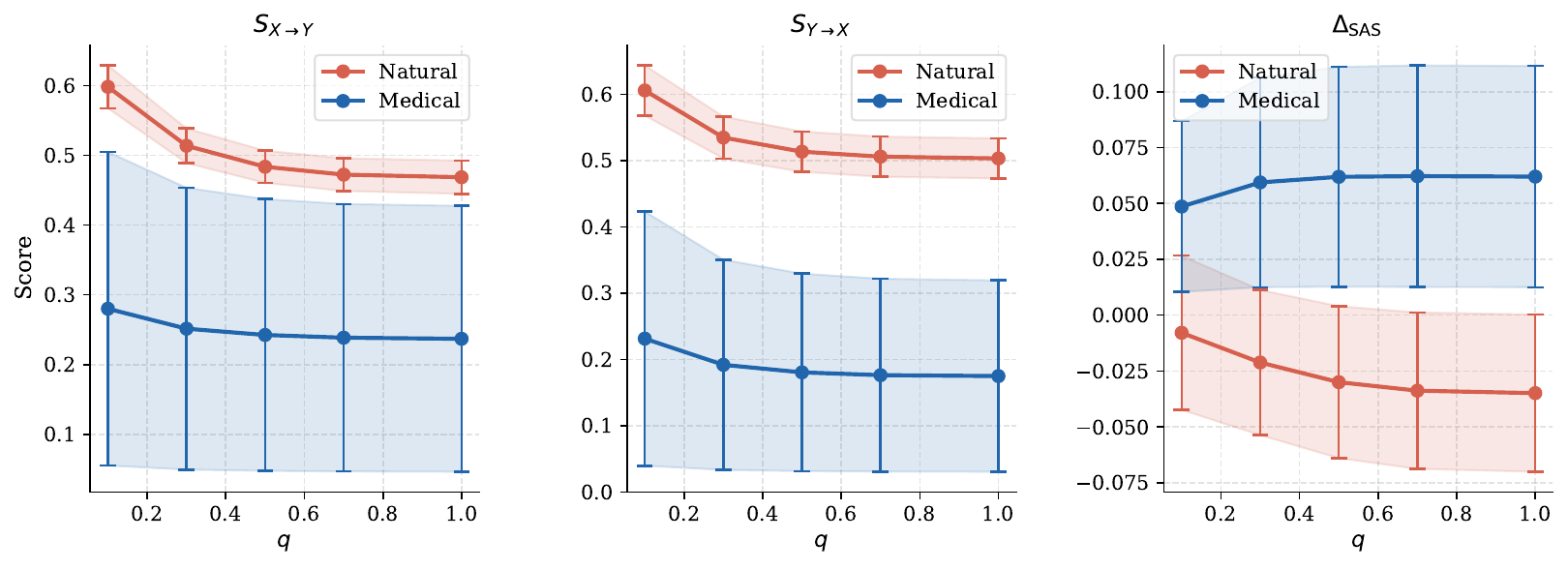}
    \caption{Sensitivity of SAS to the quantile parameter $q$.
    Each panel reports the mean score (line) and standard deviation (shaded band/error bars) across all models, averaged over Natural (blue) and Medical (red) benchmarks. 
    Natural-domain models exhibit consistently higher, more stable scores across $q$, whereas Medical-domain scores are lower with greater variance, reflecting higher heterogeneity. 
    Both directional scores decrease monotonically as $q$ increases, indicating alignment is concentrated in the top spectral decomposition. 
    The asymmetry $\Delta_{\mathrm{SAS}}$ remains near zero for Natural models and slightly positive for Medical models for $q \geq 0.3$, showing a higher bias towards image-to-text alignment.} \label{fig:sweeping_q}
\end{figure*}

Figure~\ref{fig:sweeping_q} shows the domain-level $S_{X{\to}Y}$, $S_{Y{\to}X}$  and $\Delta_{\text{SAS}}$ as a function of $q$, respectively.

The sweep confirms that $q = 0.1$ is the optimal operating point for SAS. The natural-medical gap in $S_{X{\to}Y}$ is largest at $q = 0.1$ ($+0.3180$) and decreases monotonically as more eigenmodes are included, reaching $+0.2316$ at $q = 1.0$. The decay is moderate on both sides: natural $S_{X{\to}Y}$ falls $22\%$ from $q = 0.1$ to $q = 1.0$ ($0.5981 \to 0.4685$), while medical falls $15\%$ ($0.2800 \to 0.2369$). The primary differentiator is therefore the absolute level of alignment, not the shape of spectral concentration, and the top-decile eigenmodes provide the cleanest diagnostic window. A similar pattern is shown for $S_{Y \to X}$, but with a larger gap across domains. 

The imbalance sweep reveals additional spectral structure. Medical $\Delta_{\text{SAS}}$ grows from $+0.0486$ at $q = 0.1$ to $+0.0622$ at $q = 0.7$, after which it stabilises. Natural $\Delta_{\text{SAS}}$ drifts monotonically from $-0.0079$ toward $-0.035$. This pattern indicates that
increasing $q$ makes the text modality less recoverable.


\subsection{Exp. 4 — SAS Is the Strongest Predictor of Retrieval in the Medical Domain}

\subsubsection{Design}

We ask which metric serves as the best zero-label proxy for retrieval performance, and whether directional SAS predicts its matching retrieval direction more strongly than the opposite direction. 
For each of the 52 \texttt{(model $\times$ dataset)} pairs, we compute the full metric suite alongside R@1 both directions, and \textit{rSum} as the sum of R@1, R@5, and R@10 both directions. Spearman rank correlations are computed domain-stratified (natural vs.\ medical). 

\subsubsection{Results}

\begin{table}[t]
\centering
\caption{Dataset-level retrieval averages ($\mu \pm \sigma$ over 13 models).}
\label{tab:exp4_retrieval_means}
\begin{tabular}{llcc}
\toprule
& & \multicolumn{2}{c}{\textbf{R@1}} \\
\cmidrule(lr){3-4}
\textbf{Dataset} & \textbf{Domain} & $I{\to}T$ & $T{\to}I$ \\
\midrule
MS-COCO 2014 & Natural & $0.189 \pm 0.020$ & $0.138 \pm 0.022$ \\
Flickr30k & Natural & $0.873 \pm 0.056$ & $0.717 \pm 0.076$ \\
\midrule
ROCO      & Medical & $0.041 \pm 0.031$ & $0.045 \pm 0.031$ \\
MIMIC-CXR & Medical & $0.000 \pm 0.000$ & $0.000 \pm 0.000$ \\
\bottomrule
\end{tabular}
\end{table}

\begin{table}[htbp]
\centering
\caption{Domain-stratified Spearman $\rho$ correlation with retrieval.}
\label{tab:exp4_spearman}
\small
\begin{tabular}{lcc}
\toprule
\textbf{Metric} & $\boldsymbol{\rho}_{\text{nat}}$ & $\boldsymbol{\rho}_{\text{med}}$ \\ 
\midrule
$S_{Y \to X}$   & \cNegM{-0.439}$^{*}$   & \cPosS{+0.954}$^{***}$ \\ 
SAS             & \cNegM{-0.540}$^{**}$  & \cPosS{+0.969}$^{***}$ \\ 
$S_{X \to Y}$   & \cNegM{-0.531}$^{**}$  & \cPosS{+0.965}$^{***}$ \\ 
CKA             & \cNegM{-0.426}$^{*}$   & \cPosS{+0.911}$^{***}$ \\ 
CORAL           & \cNeu{+0.085}         & \cPosM{+0.640}$^{***}$ \\ 
SVCCA           & \cNegS{-0.683}$^{***}$ & \cPosS{+0.941}$^{***}$ \\ 
MMD             & \cNeu{-0.042}         & \cNegM{-0.248}         \\ 
$\Delta_{\text{SAS}}$ & \cNeu{+0.074}   & \cPosS{+0.789}$^{***}$ \\ 
RMG             & \cNegM{-0.259}         & \cPosM{+0.228}         \\ 
\bottomrule
\end{tabular}
\end{table}

Table~\ref{tab:exp4_spearman} shows the domain-stratified correlations. Within the medical domain, all SAS variants reach $\rho_{\text{med}}$ above $+0.950$, with SAS the strongest at $+0.969$***. CKA and SVCCA also perform well ($+0.911$*** and $+0.941$*** respectively), while CORAL is significant but weaker ($+0.640$***). Within the natural domain, all of these metrics turn negative or non-significant, 
with SVCCA the most affected ($-0.683$***) and CORAL the least ($+0.085$). MMD is non-significant in both domains ($-0.042$ natural, $-0.248$ medical), suggesting it responds to the modality gap rather than within-domain model quality. $\Delta_{\text{SAS}}$ is non-significant within natural ($+0.074$) but moderately strong within medical ($+0.789$***), consistent with imbalance being a medical-domain phenomenon. RMG is non-significant in both settings, as in Experiment~1.

\subsection{Benchmarking with Medical CLIP Versions}

\subsubsection{Design}

We benchmark whether medical VLMs show alignment on ROCO and MIMIC-CXR.
We consider the PubMedCLIP and PLIP medical VLMs. We acknowledge the low data scale they were trained on.
We show R@1 and representation alignment metrics for each \texttt{(model x dataset)} pair. 

\subsubsection{Results}

\begin{table}[t]
\centering
\caption{Summary of representation alignment metrics for medical models on ROCO and MIMIC-CXR. 
}\label{tab:exp6_alignment}
\begin{tabular}{lrrrr}
\toprule
& \multicolumn{2}{c}{\textbf{PLIP}} & \multicolumn{2}{c}{\textbf{PubMedCLIP}} \\
\cmidrule(lr){2-3} \cmidrule(lr){4-5}
\textbf{Metric} & \textbf{ROCO} & \textbf{MIMIC-CXR} & \textbf{ROCO} & \textbf{MIMIC-CXR} \\
\midrule
CORAL                   & $0.5564$ &  $0.0531$ & $0.0670$ & $0.0305$ \\
CKA                   & $0.181$ & $0.024$ & $0.623$ & $0.034$ \\
SVCCA                 & $0.315$ & $0.089$ & $0.640$ & $0.120$ \\
RMG                   & $0.691$ & $0.682$ & $0.732$ & $0.721$ \\
MMD                   & $0.870$ & $1.014$ & $0.953$ & $1.230$ \\
$S_{X{\to}Y}$         & $0.303$ & $0.023$ & $0.628$ & $0.076$ \\
$S_{Y{\to}X}$         & $0.202$ & $0.022$ & $0.598$ & $0.059$ \\
$\Delta_{\text{SAS}}$ & $+0.100$ & $+0.001$ & $+0.029$ & $+0.018$ \\
SAS & $0.2523$ & $0.0225$ & $0.612$ & $0.067$ \\

\bottomrule
\end{tabular}
\end{table}

\begin{table}[t]
\centering
\caption{Retrieval performance 
on ROCO and MIMIC-CXR}
\label{tab:exp6_retrieval}
\begin{tabular}{llrrr}
\toprule
& & \multicolumn{2}{c}{\textbf{R@1}} &  \\
\cmidrule(lr){3-4}
\textbf{Model} & \textbf{Dataset} & $I{\to}T$ & $T{\to}I$ & \textbf{rSum} \\
\midrule
PLIP        & ROCO & $0.002$ & $0.004$ & $5.7$   \\
PubMedCLIP & ROCO & $0.047$ & $0.053$ & $84.9$ \\
\midrule
PLIP        & MIMIC-CXR & $0.000$ & $0.000$ & $0.1$  \\
PubMedCLIP & MIMIC-CXR & $0.000$ & $0.000$ & $0.2$ \\
\bottomrule
\end{tabular}
\end{table}

Table~\ref{tab:exp6_alignment} shows a divergence on ROCO: PubMedCLIP scores substantially higher on all alignment metrics: CKA $0.623$ vs $0.181$, SVCCA $0.640$ vs $0.315$, and both SAS directions ($S_{X \to Y}$ $0.628$ vs $0.303$, $S_{Y \to X}$ $0.598$ vs $0.202$). $\Delta_{\text{SAS}}$ of $+0.029$ indicates near-symmetric alignment, meaning the text encoder has substantially caught up to the image eigenstructure. PLIP's $\Delta_{\text{SAS}}$ of $+0.100$ displays the opposite, with the image modality dominating, reflecting a lack of text encoder adaptation. RMG shows little separation between the two models ($0.691$ vs $0.732$), consistent with its insensitivity observed in earlier experiments.
%
Table~\ref{tab:exp6_retrieval} confirms that neither specialist achieves viable retrieval on MIMIC-CXR, demonstrating that broad medical pretraining alone is insufficient for radiology-report retrieval.



\section{Instance-Level SAS for Individual Image--Caption Pairs}
\label{subsec:local-sas}

Since SAS is a global metric, it summarizes alignment over an entire dataset, but does not assess individual pairs.
We extend SAS to a \emph{local} formulation that reuses the spectral prior fitted
on a reference population and scores a single pair at inference time. This involves a two-stage procedure. 
First, given population representations $X \in \mathbb{R}^{n \times d}$ and
$Y \in \mathbb{R}^{n \times d}$, the \emph{fit} stage computes the standard
SAS quantities: eigenbasis $V$, eigenvalues $\Lambda$,
per-mode text variance $\sigma^2_{Y,k}$, and active index set $\mathcal{I}_q$, and
freezes them as a prior.
No further updates to these quantities occur in the \emph{scoring} stage.
Second, for a new pair $(\mathbf{x}_i, \mathbf{y}_i)$, we center with the population means
$\boldsymbol{\mu}_X$ and $\boldsymbol{\mu}_Y$,
project into the frozen eigenbasis,
\begin{equation}
    z^x_{ik} = (\mathbf{x}_i - \boldsymbol{\mu}_X)^\top \mathbf{v}_k,
    \qquad
    z^y_{ik} = (\mathbf{y}_i - \boldsymbol{\mu}_Y)^\top \mathbf{v}_k,
\end{equation}
and form the rank-1 cross-covariance diagonal
$a^{(i)}_k = z^x_{ik}\,z^y_{ik}$,
which is the contribution of pair $i$ to the full matrix $A$.
Substituting into Eq.(~\ref{eq:sas_corrcoef}) with the frozen normalization scale yields the
instance-level normalized correlation,
\begin{equation}
    \rho^{(i)}_k = \frac{a^{(i)}_k}{\sqrt{\lambda_k\,\sigma^2_{Y,k} + \varepsilon}},
\end{equation}
and the local directional score is
\begin{equation}
    S^{(i)}_{X \to Y}
    = \frac{\displaystyle\sum_{k \in \mathcal{I}_q} \lambda_k \,|\rho^{(i)}_k|}
           {\displaystyle\sum_{k \in \mathcal{I}_q} \lambda_k}.
    \label{eq:local-sas}
\end{equation}
Note that $S^{(i)}_{X \to Y}$ is on the same scale as the population score
$S_{X \to Y}$: a value above (below) $S_{X \to Y}$ indicates that pair $i$
is more (less) aligned than the dataset average in the dominant image directions.
Unlike the population score, $S^{(i)}_{X \to Y}$ is unbounded above~1 because
a single pair's product $z^x_{ik} z^y_{ik}$ can exceed its expected scale
$\sqrt{\lambda_k \sigma^2_{Y,k}}$.

\subsection{Application to Dental Panoramic Radiographs}

\begin{figure*}[t]
    \centering
\includegraphics[width=1\linewidth]{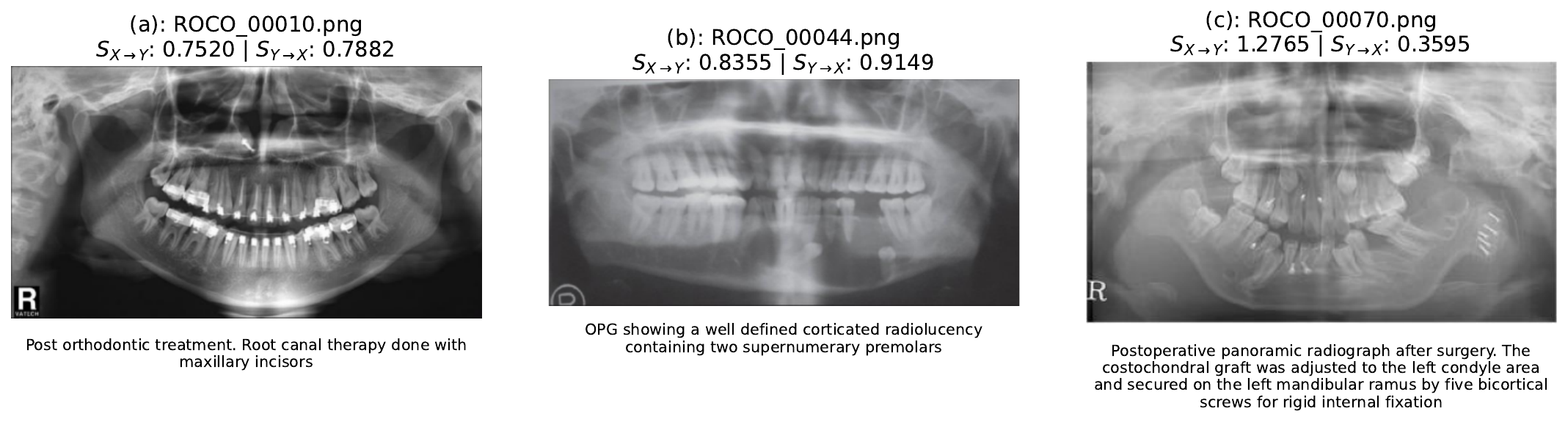}
      \caption{%
        Instance-level SAS scores for three OPGs from the
        ROCO dataset~\cite{pelka2018roco}, encoded by CLIP-Large.
        Each panel shows the radiograph together with its original radiology report and the two directional scores $S^{(i)}_{X \to Y}$ (image to
        text) and $S^{(i)}_{Y \to X}$ (text to image), computed via
        Eq.~\eqref{eq:local-sas} against a spectral prior fitted on the full
        ROCO test split.
        A score above the global $S_{X \to Y}$ indicates stronger-than-average
        cross-modal alignment in the dominant image directions.
        From left to right: (\textbf{a})~post-orthodontic OPG with root canal
        therapy on maxillary incisors; (\textbf{b})~OPG revealing a
        corticated radiolucency enclosing two supernumerary premolars;
        (\textbf{c})~postoperative panoramic radiograph following costochondral
        graft fixation with five bicortical screws.
    } \label{fig:dental_sas}
\end{figure*}

We perform a case study in which we instantiate the instance-level SAS on three orthopantomograms (OPGs) from the ROCO
dataset~\cite{pelka2018roco}, each paired with its original radiology caption.
An OPG is a panoramic dental X-ray that captures a wide, single-image view of all the upper and lower teeth, jaws, and surrounding structures.
In Figure~\ref{fig:dental_sas}, we show three OPG with relative local asymmetric SAS scores. 
The spectral global priors are calculated on CLIP-Large embeddings of the full ROCO test split, resulting to values of $S_{X \to Y}$=0.9045 and $S_{Y \to X}$=0.3664.
Panel~(a) shows a post-orthodontic OPG with root canal therapy on the maxillary 
incisors. Both directional scores ($S_{X\to Y}^{(i)} = 0.752$, $S_{Y\to X}^{(i)} 
= 0.789$) are below the global $S_{X\to Y}$ but well above $S_{Y\to X}$, indicating 
moderate, near-symmetric alignment for this pair.
Panel~(b) shows an OPG with a corticated radiolucency enclosing two supernumerary 
premolars. Scores are $S_{X\to Y}^{(i)} = 0.836$ and $S_{Y\to X}^{(i)} = 0.915$,  close to the global $S_{X\to Y}$ and significantly higher to the global $S_{Y\to X}$.
specific terminology that anchors the dominant image eigenmodes.
Panel~(c) shows a postoperative panoramic radiograph after costochondral graft 
fixation with five bicortical screws. The image contains surgical hardware and 
altered bone geometry that the paired report describes only at a high level. This is 
reflected in $S_{X\to Y}^{(i)} = 1.277$ and $S_{Y\to X}^{(i)} = 0.360$: the 
image-to-text score exceeds even the global $S_{X\to Y}$, while the reverse 
direction sits near the global $S_{Y\to X}$, indicating that the report may only capture a subset of the structural information present in the scan. 

\section{Discussion}
Nine of ten metrics detect alignment degradation when moving from natural
to medical data. 
The additional finding is that this degradation is not symmetric across
modalities. Across all 13 general-purpose VLMs and both medical datasets,
$\Delta_{\mathrm{SAS}}$ is consistently positive, while it is statistically
not different from zero on natural data. This directional asymmetry is
not captured by CKA, SVCCA, CORAL, MMD, or RMG, since all of these metrics
aggregate the two anchor choices into a single scalar and discard directional information by construction.

A positive $\Delta_{\mathrm{SAS}}$ means that the dominant principal
directions of the image representations are more strongly recovered from the text
representations than the reverse. 
We argue that clinical reports may contain domain-specific linguistic patterns (severity qualifiers, anatomical templates, etc.) that have no consistent visual grounding that, in turn, the image encoder, trained on natural data, does not generate representations that recover this text-side variation. 
However, these experiments do not clearly disambiguate these explanations: controlled
pretraining interventions would be needed to do so. $\Delta_{\mathrm{SAS}}$
identifies which modality is the bottleneck, while determining the cause requires further experimentation.

We further argue that SAS metrics and cosine similarity can be useful together:
in particular, $\Delta_{\cos}$ measures the margin between matched and unmatched pair similarity scores, which is the direct condition for robust retrieval performance, 
while, $\Delta_{\mathrm{SAS}}$ measures directional eigenspace recoverability and identifies which encoder is the weaker one. In practice, if $\Delta_{\cos}$ is low and $\Delta_{\mathrm{SAS}} \approx 0$, alignment has collapsed symmetrically and the standard intervention may be joint fine-tuning on in-domain paired data. Conversely, if $\Delta_{\cos}$ is low and $\Delta_{\mathrm{SAS}}$ is substantially positive, the image encoder retains more recoverable structure than the text encoder, suggesting that text-side adaptation such as domain-specific language model pretraining or text-encoder fine-tuning on clinical reports may be a better intervention. 

Our work is not without limitations. 
First, SAS, in its current formulation, requires equal feature dimensionality ($d_1 = d_2 = d$), limiting its applicability to models whose vision and language encoders share an identical output dimension. 
Second, despite strong alignment scores, PubMedCLIP and PLIP both achieve near-zero retrieval performance on MIMIC-CXR ($\text{rSum} \leq 0.2$), demonstrating that alignment in representation space is a necessary but not sufficient condition for downstream retrieval. This gap suggests that the distribution of MIMIC-CXR reports, with jargon-heavy clinical notes, constitutes a harder distributional challenge than the metrics alone can resolve. 
Third, within the natural domain, alignment metrics exhibit negative or non-significant correlations with retrieval ($\rho_\text{nat}$ ranging from $-0.683$ to $+0.085$), an artefact of model-rank reversal between MSCOCO and Flickr30k rather than metric failure. 
Fourth, the sensitivity of SAS depends on the calibration of the eigenspectrum depth parameter $q$. As demonstrated in our experiments, expanding $q$ toward the full spectrum ($q = 1.0$) introduces lower eigenmodes that dilute the absolute alignment gap and introduce noisy, uninformative directional drift, requiring careful, domain-specific selection of the active quantile. 
Fifth, while SAS is lightweight at inference, computing full eigenspectra for very high-dimensional embeddings introduces moderate overhead that may require optimization for real-time applications. 

Future work may explore several directions. We plan to expand on more datasets in both domains and fine-tune models to assess how domain adaptation can reduce information imbalance. 
Furthermore, we plan to stratify by specific medical domain, \textit{e.g.}, dentistry, orthopedics, gastroenterology, etc, to measure whether different fields carry different patterns. Furthermore, several human-computer interaction (HCI) experiments would be needed to perform post-hoc evaluations, and also human signals could be included in either pre-training or fine-tuning to guide optimization for better medical representation learning, penalizing directional imbalance to produce better-calibrated medical VLMs. 
In particular, grounding the per-eigenmode profile $\{\rho_k\}$ in human-readable concepts might open the door to HCI tools designed for clinical settings, where radiologists could directly inspect which visual features lack a textual counterpart and provide targeted feedback to guide model improvement. 
 Overall, SAS adds a missing diagnostic dimension to multimodal evaluation, which is not meant to replace but to be used in synergy with current evaluation metrics.

\section{Conclusion}

General-purpose VLMs fall short in the medical domain, and existing alignment metrics cannot directly explain why. 
%
Decomposing cross-modal alignment into directed, spectral scores such as SAS sheds light on the directional modality imbalance underlying this failure.
By benchmarking a comprehensive suite of representation alignment metrics across natural and medical image-text pairs, this work provides a label-free framework that detects not only that alignment degrades under domain shift, but which modality drives that degradation. 
The evidence confirms that medical images carry richer structural information than their paired clinical reports can recover, a finding with direct implications for how retrieval systems should be designed and evaluated in healthcare settings. 
Beyond retrieval, we see this framework as useful toward more explainable vision-language systems, where per-eigenmode alignment profiles could be presented to clinicians to reveal which visual concepts lack a reliable textual counterpart. 
We hope this work encourages the information retrieval, HCI, and medical AI communities to treat cross-modal alignment as a transparent, interpretable, and actively improvable property of clinical AI.

\bibliographystyle{ACM-Reference-Format}
\bibliography{vlms}

@article{liang2022mind,
  title={Mind the gap: Understanding the modality gap in multi-modal contrastive representation learning},
  author={Liang, Victor Weixin and Zhang, Yuhui and Kwon, Yongchan and Yeung, Serena and Zou, James Y},
  journal={Advances in Neural Information Processing Systems},
  volume={35},
  pages={17612--17625},
  year={2022}
}

@inproceedings{schrodi2025TwoEO,
  title={Two Effects, One Trigger: On the Modality Gap, Object Bias, and Information Imbalance in Contrastive Vision-Language Models},
  author={Simon Schrodi and David T. Hoffmann and Max Argus and Volker Fischer and Thomas Brox},
  booktitle={ICLR},
  year={2025}
}

@inproceedings{wang2020understanding,
  title={Understanding contrastive representation learning through alignment and uniformity on the hypersphere},
  author={Wang, Tongzhou and Isola, Phillip},
  booktitle={International Conference on Machine Learning},
  pages={9929--9939},
  year={2020},
  organization={PMLR}
}

@inproceedings{radford2021learning,
  title={Learning Transferable Visual Models From Natural Language Supervision},
  author={Radford, Alec and Kim, Jong Wook and Hallacy, Chris and Ramesh, Aditya and Goh, Gabriel and Agarwal, Sandhini and Sastry, Girish and Askell, Amanda and Mishkin, Pamela and Clark, Jack and Krueger, Gretchen and Sutskever, Ilya},
  booktitle={International Conference on Machine Learning (ICML)},
  pages={8748--8763},
  year={2021}
}

@inproceedings{zhai2023sigmoid,
  title={Sigmoid Loss for Language Image Pre-Training},
  author={Zhai, Xiaohua and Mustafa, Basil and Kolesnikov, Alexander and Beyer, Lucas},
  booktitle={Proceedings of the IEEE/CVF International Conference on Computer Vision (ICCV)},
  pages={11975--11986},
  year={2023}
}

@inproceedings{xu2024demystifying,
  title={Demystifying CLIP Data},
  author={Xu, Hu and Xie, Saining and Tan, Xiaoqing Ellen and Huang, Po-Yao and Howes, Russell and Sharma, Vasu and Li, Shang-Wen and Ghosh, Gargi and Zettlemoyer, Luke and Feichtenhofer, Christoph},
  booktitle={International Conference on Learning Representations (ICLR)},
  year={2024}
}

@article{tschannen2025siglip2,
  title={SigLIP 2: Multilingual Vision-Language Encoders with Improved Semantic Understanding, Localization, and Dense Features},
  author={Tschannen, Michael and Alabdulmohsin, Ibrahim and Wang, Xiao and Steiner, Andreas and Zhai, Xiaohua and Beyer, Lucas and Kolesnikov, Alexander},
  journal={arXiv preprint arXiv:2502.14786},
  year={2025}
}

@article{oord2018representation,
  title={Representation learning with contrastive predictive coding},
  author={Oord, Aaron van den and Li, Yazhe and Vinyals, Oriol},
  journal={arXiv preprint arXiv:1807.03748},
  year={2018}
}

@inproceedings{jia2021scaling,
  title={Scaling up visual and vision-language representation learning with noisy text supervision},
  author={Jia, Chao and Yang, Yinfei and Xia, Ye and Chen, Yi-Ting and Parekh, Zarana and Pham, Hieu and Le, Quoc and Sung, Yun-Hsuan and Li, Zhen and Duerig, Tom},
  booktitle={International conference on machine learning},
  pages={4904--4916},
  year={2021},
  organization={PMLR}
}

@article{chuang2025meta,
  title={Meta clip 2: A worldwide scaling recipe},
  author={Chuang, Yung-Sung and Li, Yang and Wang, Dong and Yeh, Ching-Feng and Lyu, Kehan and Raghavendra, Ramya and Glass, James and Huang, Lifei and Weston, Jason and Zettlemoyer, Luke and others},
  journal={arXiv preprint arXiv:2507.22062},
  year={2025}
}

@inproceedings{schuhmann2022laionb,
  title={{LAION}-5B: An open large-scale dataset for training next generation image-text models},
  author={Christoph Schuhmann and
          Romain Beaumont and
          Richard Vencu and
          Cade W Gordon and
          Ross Wightman and
          Mehdi Cherti and
          Theo Coombes and
          Aarush Katta and
          Clayton Mullis and
          Mitchell Wortsman and
          Patrick Schramowski and
          Srivatsa R Kundurthy and
          Katherine Crowson and
          Ludwig Schmidt and
          Robert Kaczmarczyk and
          Jenia Jitsev},
  booktitle={Thirty-sixth Conference on Neural Information Processing Systems Datasets and Benchmarks Track},
  year={2022},
  url={https://openreview.net/forum?id=M3Y74vmsMcY}
}

@inproceedings{cherti2023openclip,
    author    = {Cherti, Mehdi and Beaumont, Romain and Wightman, Ross and Wortsman, Mitchell and Gabriel, Luke and Dehghani, Mostafa and Schmidt, Ludwig and Jitsev, Jenia and Schuhmann, Christoph and Kaczmarczyk, Robert},
    title     = {Reproducible Scaling Laws for Contrastive Language-Image Learning},
    booktitle = {Proceedings of the IEEE/CVF Conference on Computer Vision and Pattern Recognition (CVPR)},
    month     = {June},
    year      = {2023},
    pages     = {2818-2829}
}

@inproceedings{sun2016deep,
  title={Deep CORAL: Correlation Alignment for Deep Domain Adaptation},
  author={Sun, Baochen and Saenko, Kate},
  booktitle={Computer Vision--ECCV 2016 Workshops: Amsterdam, The Netherlands, October 8-10, 16, 2016, Proceedings, Part III},
  pages={443--450},
  year={2016},
  organization={Springer}
}

@inproceedings{kornblith2019similarity,
  title={Similarity of neural network representations revisited},
  author={Kornblith, Simon and Norouzi, Mohammad and Lee, Honglak and Hinton, Geoffrey},
  booktitle={International Conference on Machine Learning},
  pages={3519--3529},
  year={2019},
  organization={PMlR}
}

@inproceedings{sun2016return,
  title={Return of frustratingly easy domain adaptation},
  author={Sun, Baochen and Feng, Jiashi and Saenko, Kate},
  booktitle={Proceedings of the AAAI conference on artificial intelligence},
  volume={30},
  number={1},
  year={2016}
}

@inproceedings{karpathy2015deep,
  title={Deep visual-semantic alignments for generating image descriptions},
  author={Karpathy, Andrej and Fei-Fei, Li},
  booktitle={Proceedings of the IEEE Conference on Computer Vision and Pattern Recognition},
  pages={3128--3137},
  year={2015}
}

@article{raghu2017svcca,
  title={Svcca: Singular vector canonical correlation analysis for deep learning dynamics and interpretability},
  author={Raghu, Maithra and Gilmer, Justin and Yosinski, Jason and Sohl-Dickstein, Jascha},
  journal={Advances in Neural Information Processing Systems},
  volume={30},
  year={2017}
}

@article{gretton2012kernel,
  title={A kernel two-sample test},
  author={Gretton, Arthur and Borgwardt, Karsten M and Rasch, Malte J and Sch{\"o}lkopf, Bernhard and Smola, Alexander},
  journal={The Journal of Machine Learning Research},
  volume={13},
  number={1},
  pages={723--773},
  year={2012},
  publisher={JMLR. org}
}

@inproceedings{li2022blip,
  title={Blip: Bootstrapping language-image pre-training for unified vision-language understanding and generation},
  author={Li, Junnan and Li, Dongxu and Xiong, Caiming and Hoi, Steven},
  booktitle={International conference on machine learning},
  pages={12888--12900},
  year={2022},
  organization={PMLR}
}

@article{liu2023visual,
  title={Visual instruction tuning},
  author={Liu, Haotian and Li, Chunyuan and Wu, Qingyang and Lee, Yong Jae},
  journal={Advances in neural information processing systems},
  volume={36},
  pages={34892--34916},
  year={2023}
}

@article{alayrac2022flamingo,
  title={Flamingo: a visual language model for few-shot learning},
  author={Alayrac, Jean-Baptiste and Donahue, Jeff and Luc, Pauline and Miech, Antoine and Barr, Iain and Hasson, Yana and Lenc, Karel and Mensch, Arthur and Millican, Katherine and Reynolds, Malcolm and others},
  journal={Advances in neural information processing systems},
  volume={35},
  pages={23716--23736},
  year={2022}
}

@inproceedings{pelka2018roco,
  author    = {Pelka, Obioma and Koitka, Sven and R{\"u}ckert, Johannes and Nensa, Felix and Friedrich, Christoph M.},
  title     = {Radiology Objects in Context (ROCO): A Multimodal Image Dataset},
  booktitle = {Intravascular Imaging and Computer Assisted Stenting and Large-Scale Annotation of Biomedical Data and Expert Label Synthesis},
  year      = {2018},
  pages     = {180--189},
  publisher = {Springer},
  doi       = {10.1007/978-3-030-01364-6_19}
}

@article{johnson2019mimic,
  author  = {Johnson, Alistair E. W. and Pollard, Tom J. and Berkowitz, Seth J. and Greenbaum, Nathaniel R. and Lungren, Matthew P. and Deng, Chih-ying and Mark, Roger G. and Horng, Steven},
  title   = {MIMIC-CXR, a de-identified publicly available database of chest radiographs with free-text reports},
  journal = {Scientific Data},
  year    = {2019},
  volume  = {6},
  number  = {1},
  pages   = {317},
  doi     = {10.1038/s41597-019-0322-0}
}

@article{eslami2021does,
  title={Does clip benefit visual question answering in the medical domain as much as it does in the general domain?},
  author={Eslami, Sedigheh and De Melo, Gerard and Meinel, Christoph},
  journal={arXiv preprint arXiv:2112.13906},
  year={2021}
}

@article{huang2023visual,
    title={A visual--language foundation model for pathology image analysis using medical Twitter},
    author={Huang, Zhi and Bianchi, Federico and Yuksekgonul, Mert and Montine, Thomas J and Zou, James},
    journal={Nature Medicine},
    pages={1--10},
    year={2023},
    publisher={Nature Publishing Group US New York}
}

@inproceedings{sendak2020human,
  title={" The human body is a black box" supporting clinical decision-making with deep learning},
  author={Sendak, Mark and Elish, Madeleine Clare and Gao, Michael and Futoma, Joseph and Ratliff, William and Nichols, Marshall and Bedoya, Armando and Balu, Suresh and O'Brien, Cara},
  booktitle={Proceedings of the 2020 Conference on Fairness, Accountability, and Transparency},
  pages={99--109},
  year={2020}
}

@article{longoni2019resistance,
  title={Resistance to medical artificial intelligence},
  author={Longoni, Chiara and Bonezzi, Andrea and Morewedge, Carey K.},
  journal={Journal of Consumer Research},
  volume={46},
  number={4},
  pages={629--650},
  year={2019}
}

@article{zhang2023ethics,
  title={Ethics and governance of trustworthy medical artificial intelligence},
  author={Zhang, Jie and Zhang, Zong-ming},
  journal={BMC Medical Informatics and Decision Making},
  volume={23},
  number={1},
  pages={7},
  year={2023},
  publisher={Springer}
}

@article{ghassemi2021false,
  title={The false hope of current approaches to explainable artificial intelligence in health care},
  author={Ghassemi, Marzyeh and Oakden-Rayner, Luke and Beam, Andrew L},
  journal={The Lancet Digital Health},
  volume={3},
  number={11},
  pages={e745--e750},
  year={2021},
  publisher={Elsevier}
}

@article{hartsock2024vision,
  title={Vision-language models for medical report generation and visual question answering: A review},
  author={Hartsock, Iryna and Rasool, Ghulam},
  journal={Frontiers in artificial intelligence},
  volume={7},
  pages={1430984},
  year={2024},
  publisher={Frontiers Media SA}
}

\end{document}